\documentclass[]{alaya}
\usepackage{makecell}
\usepackage{wrapfig}
\usepackage{tabularx}
\usepackage{textcomp}
\usepackage{stfloats}
\usepackage{url}
\usepackage{verbatim}
\usepackage{titlesec}
\usepackage{tocloft}
\usepackage{adjustbox}
\usepackage{multirow}
\usepackage{pifont}
\usepackage[sc]{mathpazo}
\usepackage{tikz}
\usepackage{comment}
\usepackage{amsmath,amssymb}
\usepackage{colortbl}
\usepackage[numbers,sort&compress]{natbib}
\usepackage{color}
\usepackage{booktabs}
\usepackage{hyperref}
\usepackage{graphicx}
\usepackage{subcaption}
\RequirePackage{xspace}
\makeatletter
\DeclareRobustCommand\onedot{\futurelet\@let@token\@onedot}
\def\@onedot{\ifx\@let@token.\else.\null\fi\xspace}
\usepackage[most]{tcolorbox}
\usepackage{array}
\usepackage{siunitx}
\usepackage[table]{xcolor}
\usepackage{caption}
\definecolor{headerpurple}{HTML}{d8d2fc}
\definecolor{rowgray}{gray}{0.95}
\usepackage{CJKutf8}

\makeatother

\definecolor{adptorange}{RGB}{248, 205, 172}
\definecolor{cmpblue}{RGB}{189, 215, 238}

\definecolor{our_red}{RGB}{232,157,160}
\definecolor{our_blue}{RGB}{136,206,230}
\definecolor{our_orange}{RGB}{246,200,168}
\definecolor{our_green}{RGB}{178,211,164}

\definecolor{attn_code0}{RGB}{247,215,200}
\definecolor{attn_code1}{RGB}{238,169,139}
\definecolor{mlp_code0}{RGB}{204,201,221}
\definecolor{mlp_code1}{RGB}{102,95,153}
\definecolor{mygray}{HTML}{f0f0f0}

\definecolor{token_blue}{RGB}{84, 120, 140}

\usepackage{bbding}
\usepackage{fontawesome}
\usepackage{float}

\newlength\savewidth

\newcolumntype{x}[1]{>{\centering\arraybackslash}p{#1pt}}
\newcolumntype{y}[1]{>{\raggedright\arraybackslash}p{#1pt}}
\newcolumntype{z}[1]{>{\raggedleft\arraybackslash}p{#1pt}}

\renewcommand{\paragraph}[1]{\vspace{1.25mm}\noindent\textbf{#1}}

\usepackage{algorithm}
\usepackage{listings}

\definecolor{codeblue}{rgb}{0.25, 0.5, 0.5}
\definecolor{codekw}{rgb}{0.35, 0.35, 0.75}
\lstdefinestyle{Pytorch}{
    language = Python,
    backgroundcolor = \color{white},
    basicstyle = \fontsize{9pt}{8pt}\selectfont\ttfamily\bfseries,
    columns = fullflexible,
    aboveskip=1pt,
    belowskip=1pt,
    breaklines = true,
    captionpos = b,
    commentstyle = \color{codeblue},
    keywordstyle = \color{codekw},
}

\definecolor{green}{HTML}{009000}
\definecolor{red}{HTML}{ea4335}

\definecolor{bettergreen}{RGB}{0,128,0}

\title{HelloWorld: Enabling Socially Interactive Characters in Video World Models}
\author[1,2]{Liangyang Ouyang}
\author[2]{Ruicong Liu}
\author[2]{Xuangeng Chu}
\author[2]{Kaipeng Zhang}
\author[1]{Yoichi Sato}
\affiliation[1]{The University of Tokyo}
\affiliation[2]{Alaya Lab}
\abstract{Despite the remarkable recent progress of video world models, social interaction between users and the characters within these worlds remains unsupported. To fill this gap, we present \textbf{HelloWorld}, a video world model that enables social interaction with in-world characters. With a single button press, users can prompt the on-screen character to respond toward the camera, \textit{e.g.}, turning to the viewer, waving, nodding, or speaking a short greeting. To make these interactions natural, we propose a self-distillation pipeline that finetunes the video generation model on data synthesized by itself. Each synthesized clip contains both social interactions and camera motion, allowing the model to learn camera-pose conditioning without degrading interaction quality. At inference, we further introduce a training-free module that determines when the interaction occurs. Upon a button press, it modulates the cross-attention masks of the DiT so that the interaction-related text prompt attends only to the frames within the press window, temporally localizing the character's response. We further build \textbf{HelloWorldBench}, a 400-sample benchmark with three social interaction metrics alongside three conventional metrics, for evaluation. Experiments demonstrate that HelloWorld surpasses a variety of baselines in interaction quality, while maintaining state-of-the-art picture aesthetics and camera-pose following.}

\github{\url{https://github.com/AlayaLab/HelloWorld}}
\correspondence{\email{kaipeng.zhang@shanda.com}}
\date{\today}

\begin{document}
\maketitle

{\let\thefootnote\relax\footnotetext{This work was done during Liangyang Ouyang's internship at Alaya Lab.}}

\begin{figure}[!h]
    \centering
    \includegraphics[width=\linewidth]{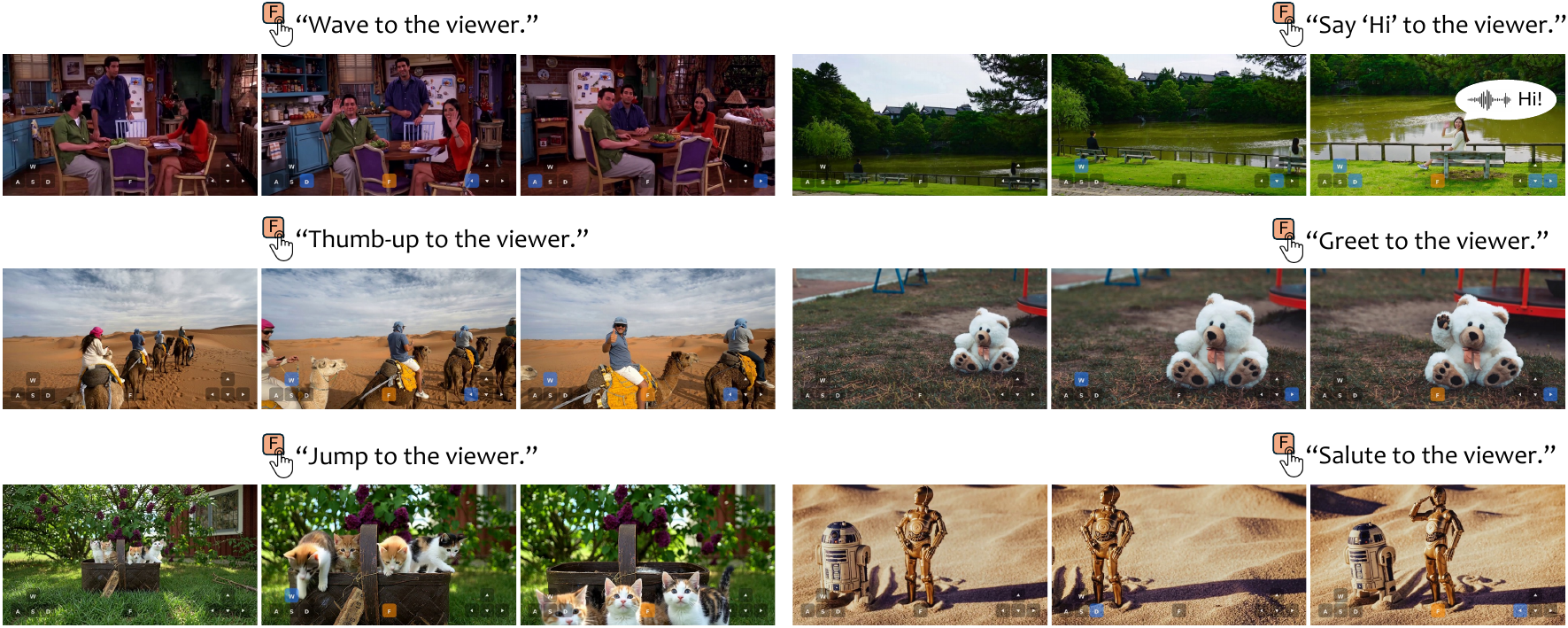}\vspace{5pt}
    \captionof{figure}{We propose HelloWorld, a video world model with socially interactive characters. With the interaction button \texttt{F}, users can prompt the on-screen character to interact with the viewer. HelloWorld supports diverse interactions across diverse characters, including humans, animals, and toys, while maintaining high-quality scene and camera-trajectory reconstruction.}
    \label{fig:teaser}
\end{figure}

\section{Introduction}

Video world models learn to dream in pixels, simulating visual worlds whose futures unfold in response to the evolving environment. Recent works make this simulation interactive, conditioning on user inputs that steer the camera through the world~\citep{sun2025worldplay} or trigger events within it~\citep{genie3, alayaworldteam2026alayaworldlonghorizonplayablevideo}. These capabilities open up applications in game production~\citep{che2025gamegen}, world simulation~\citep{videoworldsimulators2024}, and film making.

However, existing world models offer no support for social interaction between the user and the characters within these worlds. In some generated worlds, characters remain static pixels~\citep{wang2026matrix}. Some models are able to animate characters~\citep{team2026advancing}, yet their motions are ambient behaviors rather than social interactions with the user.

To fill this gap, we propose HelloWorld, an interactive video world model that enables users to actively create social interactions with in-world characters. Besides the camera trajectory and the text prompt, HelloWorld accepts an additional input, interaction button \texttt{F}. As illustrated in Fig.~\ref{fig:teaser}, when \texttt{F} is pressed, the in-world character interacts with the user, following the description in the text prompt.

HelloWorld consists of two components: \emph{self-distillation} for finetuning (training) and a \emph{temporal cross-attention mask} for inference.
The self-distillation converts a pretrained video generation model into a controllable world model using its own generations.
We first curate a set of prompts to generate videos containing both social interactions and camera motions.
For each generated video, we annotate its camera trajectory and lift the first frame into a point cloud via off-the-shelf tools.
Given the annotated trajectory, we re-render the point cloud along the camera path to synthesize a warped video, which serves as an explicit encoding of the target camera motion.
During finetuning, the warped video is injected into the model as a history condition, from which the model learns to follow the specified camera trajectory while preserving the interaction quality.
The temporal cross-attention mask is training-free and applied at inference time to control \emph{when} an interaction occurs. Specifically, upon the button \texttt{F} press, the mask modulates the cross-attention layers of the DiT such that the interaction-related text tokens attend only to the frames within the press window, thereby temporally localizing the character's response.

To evaluate interactions, we introduce HelloWorldBench, a benchmark built upon 120 high-quality images spanning diverse humans, animals, and stylized characters. Pairing these images with designed text prompts and camera trajectories yields 400 evaluation samples. Beyond standard metrics for camera accuracy and video quality, we propose three interaction-specific metrics that respectively assess \emph{what} interaction is performed, \emph{when} it occurs, and \emph{whether} it is directed toward the viewer.

Experiments on HelloWorldBench demonstrate that HelloWorld substantially outperforms existing methods on all three interaction metrics. Meanwhile, it remains competitive with or surpasses recent world models in video quality and camera following, achieving state-of-the-art performance overall. Our main contributions are summarized as:

\begin{itemize}
    \item We propose HelloWorld, an interactive video world model that enables users to actively create social interactions with in-world characters.
    \item We build HelloWorldBench, the first benchmark for social interactions in world models. It provides newly collected images, designed social prompts, camera trajectories, and a suite of interaction metrics.
    \item We conduct comprehensive evaluations, demonstrating that HelloWorld substantially outperforms existing world models in social interaction, while maintaining state-of-the-art video quality and camera-pose following.
\end{itemize}

\section{Related Work}

\subsection{Interactive Video World Models}

Video world models learn to simulate visual worlds~\citep{bai2025masks} from large-scale videos of real and virtual environments~\citep{zhou2018stereo, ling2024dl3dv, sekai2025, zhou2025omniworld, wang2026spatialvid, che2025gamegen}. Built upon video generation models~\citep{wan2025wan, yuan2026helios}, recent interactive world models~\citep{bruce2024genie, agarwal2026cosmos, feng2026matrix, zhu2025astra, gao2025longvie} further introduce controls of camera~\citep{sun2025worldplay, wang2026matrix, huang2025voyager, sun2026prisma}, keyboard~\citep{li2025hunyuan, valevski2025diffusion, decart2024oasis, guo2025mineworld, yu2025gamefactory, savva2026solaris}, skill~\citep{alayaworldteam2026alayaworldlonghorizonplayablevideo, li2026wildworld}, and event~\citep{genie3, mao2026yume1}, enabling interactive exploration of the generated world. However, while the reconstruction and exploration of the environment have been extensively studied~\citep{duan2025worldscore, li2026worldmodelbench, xu2026worldmark}, interaction with the subjects within it remains underexplored. Recent works support object interactions such as picking up and placing items~\citep{xiong2026actworld} and controlling object dynamics~\citep{yin2026holo}, and the concurrent ReactiveGWM~\citep{wang2026reactivegwm} controls interactions between NPCs, yet it is limited to a single game. Compared with these works, HelloWorld is the first video world model focusing on social interaction between in-world characters and the user, supporting diverse stylized characters across a vocabulary of social behaviors.

\subsection{Multimodal Social Interactions}

Multimodal social interaction refers to human communication across multiple modalities, including spoken language, facial expressions, gaze~\citep{liu2021generalizing, liu2024pnp, liu2024uvagaze}, gestures~\citep{cao2025socialgesture, beg2026ambigest, liu2025sfhand, liu2024single}, and body movements~\citep{muller2021multimediate, peng2026dyadit}. Earlier works focus on understanding social interactions between humans, including social video question answering~\citep{zadeh2019social, kang2025can, peng2026svbench}, social behavior classification~\citep{kim2026grasp, ouyang2025leadership}, and social conversation modeling~\citep{lee2024modeling, li2025towards, ouyang2026multi, li2026omni}. With the development of generative models, recent works turn to the generation of multi-person social interactions, including talking video generation~\citep{kong2026let, zhong2025anytalker, huang2026bind, wang2025interacthuman, wang2025fantasyportrait, ma2025playmate2, chu2026unils, peng2026actavatar, lin2026polyslgen, zhou2025evaltalker}, social interaction video generation~\citep{ouyang2026socialdirector}, and 3D motion generation~\citep{liang2024intergen, xu2024inter, ruiz2026interact2ar, yu2026socialgen}. More recently, a series of social world models~\citep{zhou2025social, yu2026building, zhang2025socioverse} employ LLM-based world models to simulate social dynamics, but they are limited to the text modality and lack multimodal interactions. Compared with these works, HelloWorld is the first to study multimodal social interaction in video world models, covering actions, gestures, facial expressions, and speech. Beyond that, HelloWorld is the first to introduce social interaction with non-human characters such as animals, robots, and toys.

\section{Proposed Method}

\subsection{Preliminary}
\label{sec:preliminary}

\paragraph{Task formulation.}
Our video world model $\mathcal{G}$ receives four inputs: a first-frame image
$\mathbf{x}^{0}$ containing the scene and its characters; a text prompt
$\mathbf{y}$ describing the world and the interaction event; a camera
trajectory $\mathcal{C}=\{\mathbf{c}^{i}\}_{i=1}^{N}$ specifying the camera movement across the $N$ frames of the video, where
$\mathbf{c}^{i}\in\mathrm{SE}(3)$ denotes the camera pose of frame $i$;
and an interaction window $\mathcal{W}=[\tau_s,\tau_e]$ specifying when
the social interaction occurs. The model generates a video
$\mathcal{V}=\{\mathbf{x}^{i}\}_{i=1}^{N}$ in which the camera follows
$\mathcal{C}$ throughout the clip, and the character interacts with the
viewer within $\mathcal{W}$.

\paragraph{Warp video condition.}
Recent work~\citep{wang2026warp} shows that a video generation model can
be finetuned into a camera-controllable world model with only lightweight
training, by conditioning on a warp video.
The warp video is a pseudo-video that re-renders the first frame along the target camera trajectory.
The warp video is fed to the DiT $\mathcal{G}$ as a history condition, providing an explicit, frame-aligned specification of the desired camera motion:
\begin{equation}
\mathcal{V}_{\mathrm{warp}}=\mathrm{warp}(\mathbf{x}^0,\mathcal{C}),
\label{eq:warp}
\end{equation}
where $\mathrm{warp}(\cdot)$ lifts $\mathbf{x}^0$ into a point cloud with
an off-the-shelf 3D reconstructor and reprojects it onto each target
camera $\mathbf{c}^i$.
Note that the warp video is inherently incomplete: regions occluded or
outside the field of view in $\mathbf{x}^0$ appear as holes after
reprojection.
It thus serves as a geometric guidance for camera motion rather than a
complete target, leaving the model to fill the missing regions.
Following this insight, HelloWorld adopts the warp video as the camera condition:
\begin{equation}
\mathcal{V}=\mathcal{G}\!\left(\boldsymbol{\epsilon};\,\mathbf{x}^{0},\,\mathbf{y},\,\mathcal{V}_{\mathrm{warp}}\right),\quad \boldsymbol{\epsilon}\sim\mathcal{N}(\mathbf{0},\mathbf{I}).
\label{eq:cond}
\end{equation}
Inside $\mathcal{G}$, the warp video is tokenized and injected as history
tokens, each sharing the temporal position embedding of its corresponding
target frame.

\subsection{Training}
\label{sec:selfdistill}

\begin{figure*}[t]
\centering
\includegraphics[width=\textwidth]{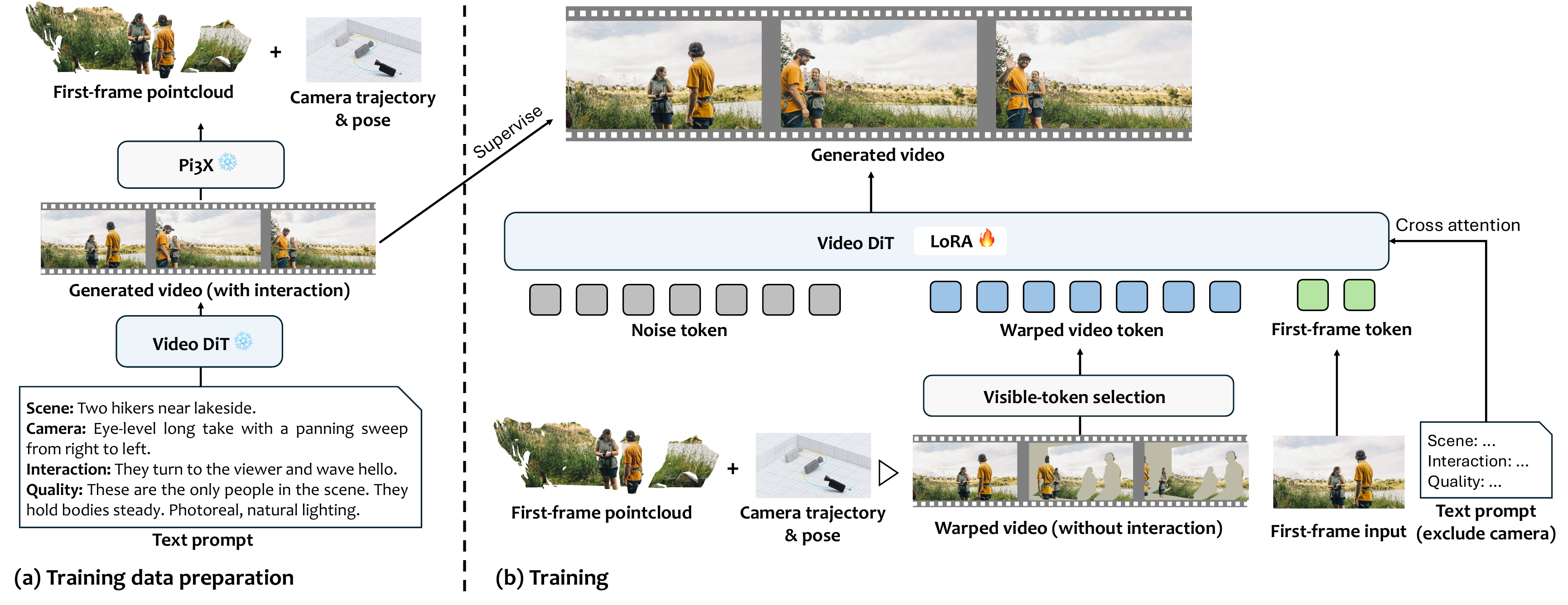}
\caption{Training pipeline of HelloWorld. (a) Training data preparation: The frozen base model generates interaction-rich clips, from which Pi3X recovers the first-frame point cloud and camera poses. (b) Training process: The point cloud is re-rendered along the trajectory into the warp video, whose visible tokens condition the DiT. A lightweight LoRA is finetuned to reconstruct the original clip, with the camera motion removed from the text prompt.}
\label{fig:train}
\end{figure*}

The base video generation model itself is capable of producing rich
social interactions~\citep{ouyang2026socialdirector}.
To preserve this ability while converting the model into a
world model, we propose a self-distillation training pipeline: the base model generates interaction-rich videos with camera motions, and is then finetuned on these self-generated videos under the warp-video
conditioning scheme in Sec.~\ref{sec:preliminary}.

First, as shown in Fig.~\ref{fig:train} (a), we prompt the base model to
synthesize training data.
The text prompt $\mathbf{y}$ contains four parts, $\mathbf{y}_{\mathrm{scene}}$, $\mathbf{y}_{\mathrm{inter}}$, $\mathbf{y}_{\mathrm{camera}}$, and $\mathbf{y}_{\mathrm{quality}}$, describing the scene, characters' interactions, camera trajectory, and video quality, respectively.
After generation, following \citet{wang2026warp}, we apply
Pi3X~\citep{wang2025pi} to recover the first-frame point cloud and the
per-frame camera trajectory.

Fig.~\ref{fig:train} (b) shows the training process.
For each clip, we render its warp video via Eq.~(\ref{eq:warp}).
A visible-token selection module then discards the warp tokens without valid source observations, \textit{i.e.}, the holes left by reprojection \cite{wang2026warp}.
The remaining warp tokens are concatenated with the noise tokens and first-frame tokens, and fed into the video DiT.
The DiT is finetuned with a lightweight LoRA under flow matching loss:
\begin{equation}
\begin{split}
\mathcal{L}=\;\mathbb{E}_{\mathbf{z}_0,\,t,\,\boldsymbol{\epsilon}}\big\|
\mathbf{v}_{\theta}\big(&\mathbf{z}_t,\,t,\,\mathbf{x}^{0},\,
\mathbf{y}_{\mathrm{scene}},\,\mathbf{y}_{\mathrm{inter}},\,\mathbf{y}_{\mathrm{quality}},\,\\
&\mathcal{V}_{\mathrm{warp}}\big)-\left(\boldsymbol{\epsilon}-\mathbf{z}_0\right)\big\|_2^2,
\end{split}
\label{eq:loss}
\end{equation}
where $\mathbf{z}_0$ is the clean latent of the training video
$\mathcal{V}$, $\mathbf{z}_t=(1-t)\,\mathbf{z}_0+t\,\boldsymbol{\epsilon}$
is its noisy interpolation at timestep $t\in[0,1]$ with
$\boldsymbol{\epsilon}\sim\mathcal{N}(\mathbf{0},\mathbf{I})$, and
$\mathbf{v}_{\theta}$ predicts the velocity field.
Notably, the camera prompt $\mathbf{y}_{\mathrm{camera}}$ is excluded to make sure that camera following is controlled solely by the warp video.
The entire self-distillation pipeline requires no external data
collection and no human annotation.

\subsection{Inference}
\label{sec:inference}

\begin{figure*}[t]
\centering
\includegraphics[width=\textwidth]{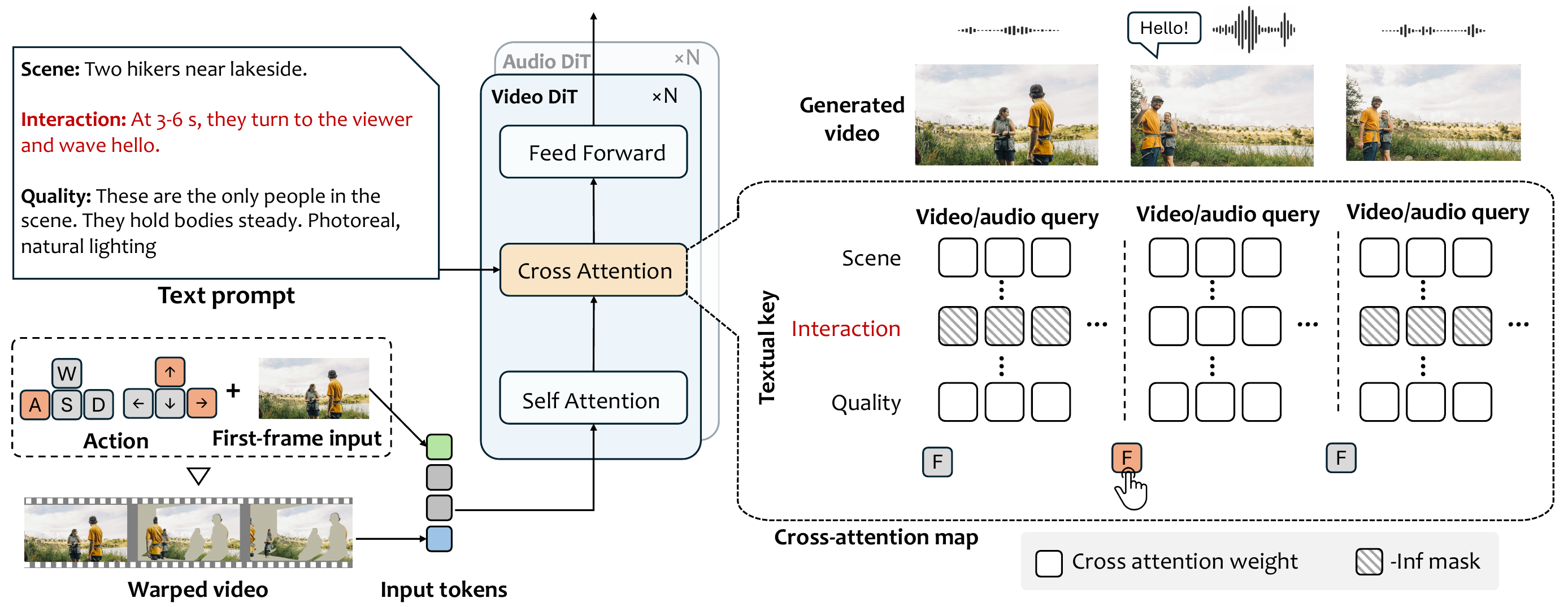}
\caption{Inference pipeline of HelloWorld.
Keyboard actions are translated into a camera trajectory and rendered
as the warp video to condition the DiT (left).
The temporal cross-attention mask blocks queries outside the \texttt{F}
press window (hatched) from attending to the interaction prompt,
temporally localizing the character's response (right).}
\vspace{-5mm}
\label{fig:inference}
\end{figure*}

Inference is illustrated in Fig.~\ref{fig:inference}. Given a first frame, a text prompt, and keyboard inputs, HelloWorld generates a video in which the character interacts with the user. The keyboard inputs are translated into a camera trajectory $\mathcal{C}$, from which the warp video is constructed via Eq.~(\ref{eq:warp}). The forward pass then follows the same procedure as training.

To further control when an interaction occurs, we propose a training-free temporal cross-attention mask.
During inference, the interaction button \texttt{F} specifies the interaction window $\mathcal{W}$: a press at time $\tau_s$ opens $\mathcal{W}=[\tau_s,\tau_e]$, spanning the subsequent seconds.
We then apply a temporal mask $M$ to the cross-attention from the audio and video tokens to the text tokens:
\begin{equation}
\begin{aligned}
M_{ij}&=
\begin{cases}
-\infty, & i\notin \mathcal{W}\; \text{and}\; j\in \mathbf{y}_{\mathrm{inter}},\\
0, & \text{otherwise},
\end{cases}\\
\mathrm{Attention}&=\mathrm{softmax}\!\left(\frac{QK^{\top}}{\sqrt{d}}+M\right)V,
\end{aligned}
\label{eq:mask}
\end{equation}
where $i$ indexes the temporal position of the query (video and audio) tokens and $j$ indexes the text tokens, with $j\in\mathbf{y}_{\mathrm{inter}}$ denoting tokens of the interaction prompt.
The mask prevents frames outside $\mathcal{W}$ from attending to
$\mathbf{y}_{\mathrm{inter}}$, so that the character responds to the
viewer precisely within the press window and behaves ambiently
elsewhere.
This temporal control is training-free and incurs negligible overhead.

\section{HelloWorldBench}

Existing video world model benchmarks~\citep{duan2025worldscore, li2026worldmodelbench, xu2026worldmark, ying2026wbench} center on scene fidelity and camera controllability, overlooking the characters within the world. We therefore propose \textbf{HelloWorldBench}, the first benchmark for evaluating viewer-directed social interactions in world models.

\subsection{Dataset Construction}

We collect 120 high-quality images from Unsplash~\citep{unsplash_website}, covering humans, animals, toys, and robots across diverse scenes and visual styles.
For each image, an LLM agent designs two to four subject-appropriate interactions, \textit{e.g.}, a human waving to the camera or a dog wagging its tail. We further design four common camera trajectories: static, scan, dolly-in, and orbit. The interaction timing is randomly assigned to early, middle, or late. Combining the images, interactions, camera trajectories, and timings yields 400 samples, spanning 264 character instances and 101 interaction types.

\begin{figure}[t]
\centering
\includegraphics[width=\columnwidth]{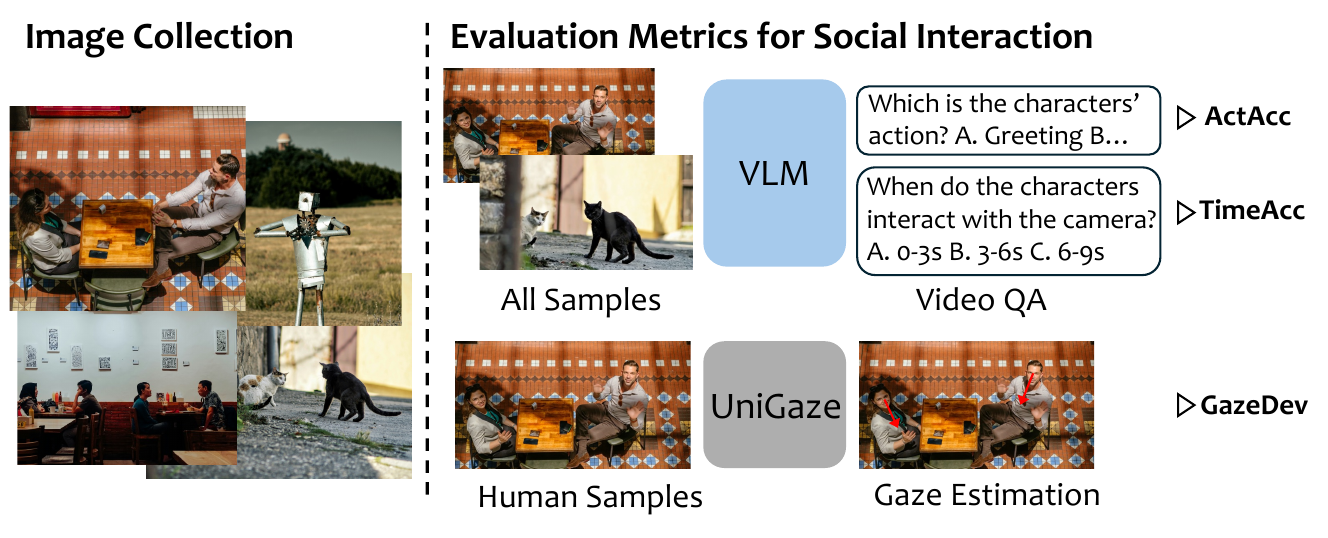}
\caption{HelloWorldBench overview. Left: we collect 120 high-quality and diverse first-frame images. Right: we use VLMs and a gaze estimator to evaluate social interactions.}
\label{fig:bench}
\end{figure}

\subsection{Evaluation Metrics}

As shown in Fig.~\ref{fig:bench}, we design three metrics decouple social interaction into \emph{what} is performed, \emph{when} it occurs, and \emph{whether} it is directed toward the viewer.

\paragraph{ActAcc\,$\uparrow$.}
ActAcc measures whether the generated character performs the prompted interaction.
We feed the generated video to a VLM judge~\citep{bai2025qwen3} with an eight-way multiple-choice question, \textit{e.g.}, ``\emph{Which is the characters' action? A.~Greeting. B.~Dancing. \dots}''.
The ground truth is the prompted action while the distractors are sampled from other interaction types in HelloWorldBench.
ActAcc is the fraction of videos for which the VLM selects the correct
answer.

\paragraph{TimeAcc\,$\uparrow$.}
TimeAcc measures whether the interaction occurs at the user-specified
moment. The VLM judge is asked ``\emph{When do the characters interact
with the camera?}'' and selects among uniformly divided three temporal
segments (\textit{e.g.}, 0--3\,s, 3--6\,s, and 6--9\,s for a 9-second clip)
and an additional \emph{no interaction} option. TimeAcc is the fraction of videos whose selected segment matches the assigned interaction window. Note that videos where the judge selects \emph{no interaction} are excluded from the calculation.

\paragraph{GazeDev\,$\downarrow$.}
GazeDev measures whether the interaction is directed toward the viewer.
Over the 217 human samples, we estimate the characters'
gaze~\citep{qin2026unigaze} and compute the mean angular deviation
between the gaze direction and the camera's optical axis within the
interaction window. A deviation of $90^{\circ}$ is assigned if no
face detected.

\paragraph{Other metrics.}
Following previous works, we also report background consistency (\textbf{BgCons}), aesthetic score (\textbf{Aesthetic}), and camera controllability (\textbf{CamCtrl}).

\section{Experiments}
\label{sec:experiments}

\subsection{Baseline Methods}
\label{sec:baselines}

We compare HelloWorld with five recent competitive video world models: WorldPlay~\citep{sun2025worldplay}, Matrix-Game 3.0~\citep{wang2026matrix}, LingBot-World~\citep{team2026advancing}, SANA-WM~\citep{zhu2026sana}, and Warp-as-History~\citep{wang2026warp}. We also include LTX-2.3~\citep{hacohen2026ltx}, the base model itself, as a pure video generation model without a trajectory interface.

\subsection{Implementation Details}
\label{sec:implementation}

Our self-distillation finetunes the base model LTX-2.3 on 156 synthesized videos for 2{,}000 steps, with a learning rate of $1\times10^{-4}$. The LoRA is rank-32, applied to all projection matrices of the self-attention layers in the video branch. The attention strength of the warp reference tokens is set to 0.3. At inference, HelloWorld and all baseline methods receive the same text prompt and camera motion input for each sample to generate 10-second videos. The resolution and frame rate of HelloWorld are set to $1280\times704$ and 24\,fps. Each sample is run with three different random seeds, reporting the averaged results. The VLM judge used in all experiments is Qwen3.6-35B-A3B~\citep{qwen36_35b_a3b}. All training and testing are conducted on a single NVIDIA H200 GPU.

\begin{table*}[t]
\centering
\caption{Comparison of HelloWorld with baseline methods on HelloWorldBench.}
\label{tab:main}
\small
\resizebox{\textwidth}{!}{
\begin{tabular}{l|c|ccc|cc|c}
\Xhline{1.0pt}
\rowcolor[gray]{0.92}
 & {\bf Camera} & \multicolumn{3}{c|}{\bf Social Interaction} & \multicolumn{2}{c|}{\bf Video Quality} & {\bf Camera} \\
\rowcolor[gray]{0.92}
\multirow{-2}{*}{\bf Method} & {\bf Input} & ActAcc $\uparrow$ & TimeAcc $\uparrow$ & GazeDev$^{\circ}$ $\downarrow$ & BgCons $\uparrow$ & Aesthetic $\uparrow$ & CamCtrl $\uparrow$ \\
\hline
\color{gray} LTX-2.3~\citep{hacohen2026ltx} & \color{gray} None & \color{gray} 42.5 & \color{gray} 52.6 & \color{gray} 38.1 & \color{gray} 94.7 & \color{gray} 5.19 & \color{gray} 31.4 \\
\hline
WorldPlay~\citep{sun2025worldplay} & Keyboard & 10.5 & 41.2 & 63.5 & 94.7 & 5.14 & 48.1 \\
Matrix-Game 3.0~\citep{wang2026matrix} & Keyboard & 8.0 & 37.5 & 77.2 & 88.3 & 4.95 & 51.1 \\
LingBot-World~\citep{team2026advancing} & Keyboard & \textbf{50.5} & 39.5 & 59.0 & 93.1 & 5.21 & 62.6 \\
SANA-WM~\citep{zhu2026sana} & Trajectory & 38.5 & 30.9 & 56.8 & 92.5 & 5.01 & 70.0 \\
Warp-as-History~\citep{wang2026warp} & Trajectory & 33.8 & 35.2 & 52.8 & 95.3 & 5.14 & 65.1 \\
\hline
\textbf{HelloWorld (ours)} & Trajectory & 41.4 & \textbf{81.7} & \textbf{40.2} & \textbf{96.9} & \textbf{5.27} & \textbf{82.9} \\
\Xhline{1.0pt}
\end{tabular}
}
\end{table*}

\subsection{Main Results}
\label{sec:results}

\paragraph{Comparison with baselines.}
As shown in Table~\ref{tab:main}, HelloWorld outperforms existing world models by a clear margin.
WorldPlay and Matrix-Game~3.0 tend to generate static scenes and thus score low across all three social interaction metrics.
LingBot-World, SANA-WM, and Warp-as-History follow the text prompt and produce character actions, LingBot-World even attains the highest ActAcc.
However, none of them controls \emph{when} the interaction occurs: their TimeAcc stays around 30\%, close to the 33.3\% random-guess level of the three-way timing question, whereas HelloWorld reaches 81.7\%.
Their larger gaze deviations further indicate that the generated characters fail to look toward the camera.
On video quality, HelloWorld performs on par with or slightly better than the best baseline, showing that self-distillation preserves the quality of the base model.
Finally, HelloWorld attains the best camera-following score, confirming that warp-video conditioning is an effective approach to converting a video generation model into a camera-controllable world model.

\begin{figure*}[t]
\centering
\includegraphics[width=\textwidth]{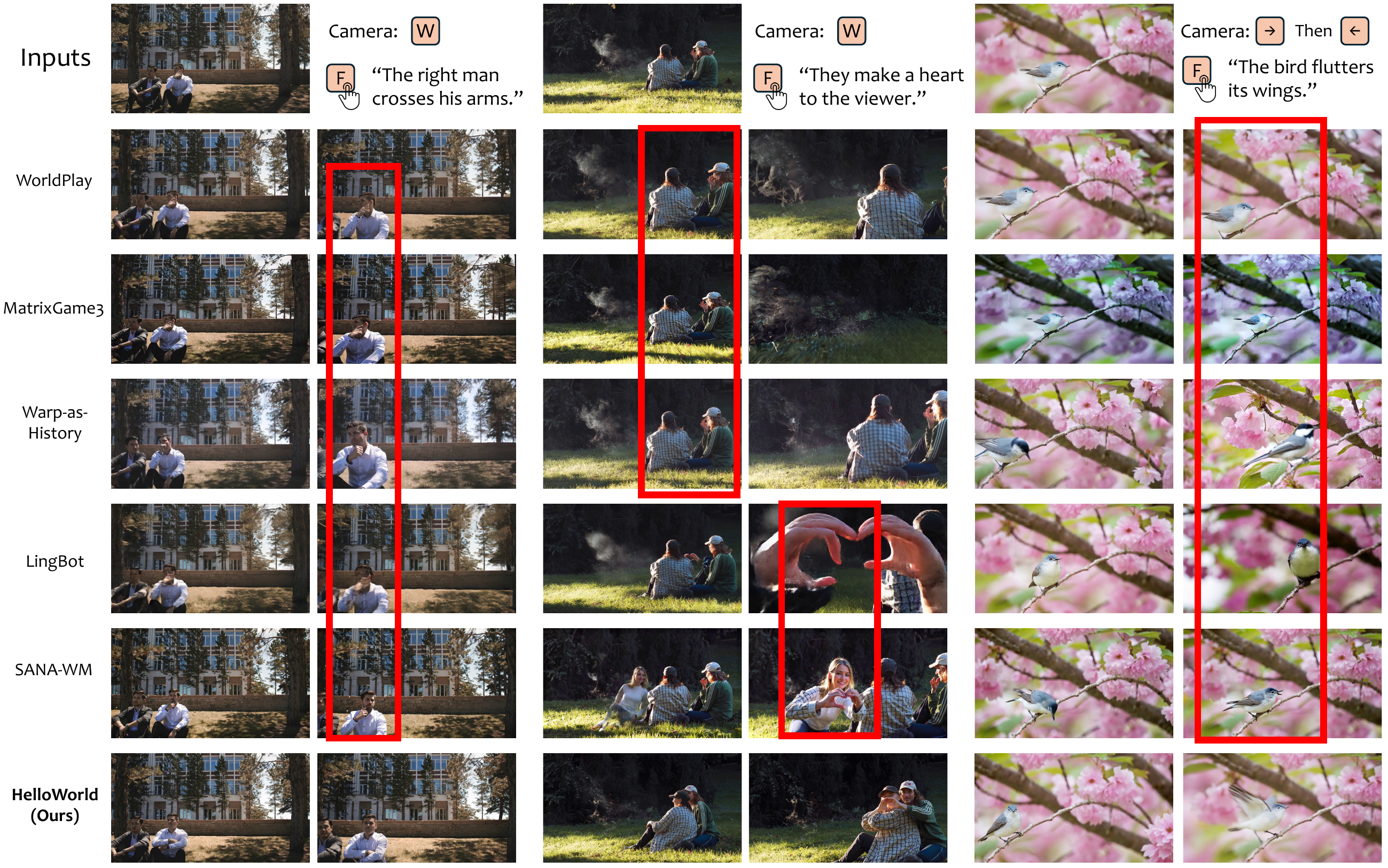}
\caption{Qualitative comparison with baseline methods.}
\label{fig:vis}
\end{figure*}

\paragraph{Visualizations.}
Fig.~\ref{fig:vis} presents the qualitative comparison with baseline methods. As highlighted by the red boxes, the baseline models often fail to generate the correct interactions: the characters neither look toward the camera nor perform the prompted actions, such as crossing arms or fluttering wings.
LingBot-World and SANA-WM do generate some actions, but are prone to
hallucination.
For ``make a heart'', an extra character abruptly appears in the scene.
In contrast, HelloWorld produces faithful viewer-directed interactions: the characters act toward the camera with vivid and natural motions, consistent with the quantitative results in Table~\ref{tab:main}.

\subsection{Ablation Studies}
\label{sec:ablations}

\begin{table*}[t]
\centering
\caption{Ablation on the training data.}
\label{tab:ablation_data}
\small
\resizebox{\textwidth}{!}{
\setlength{\tabcolsep}{5mm}
\begin{tabular}{l|ccc|cc|c}
\Xhline{1.0pt}
\rowcolor[gray]{0.92}
 & \multicolumn{3}{c|}{\bf Social Interaction} & \multicolumn{2}{c|}{\bf Video Quality} & {\bf Cam.} \\
\rowcolor[gray]{0.92}
\multirow{-2}{*}{\bf Training data} & ActAcc $\uparrow$ & TimeAcc $\uparrow$ & GazeDev$^{\circ}$ $\downarrow$ & BgCons $\uparrow$ & Aesthetic $\uparrow$ & CamCtrl $\uparrow$ \\
\hline
Real-video & 36.4 & 81.3 & 51.3 & \textbf{97.2} & 5.23 & \textbf{83.1} \\ 
Human-only & 40.4 & \textbf{82.4} & 42.3 & 96.8 & \textbf{5.27} & 82.7 \\ 
Full & \textbf{41.4} & \underline{81.7} & \textbf{40.2} & \underline{96.9} & \textbf{5.27} & \underline{82.9} \\ 
\Xhline{1.0pt}
\end{tabular}
}
\end{table*}

\begin{figure}[t]
\centering
\includegraphics[width=0.8\linewidth]{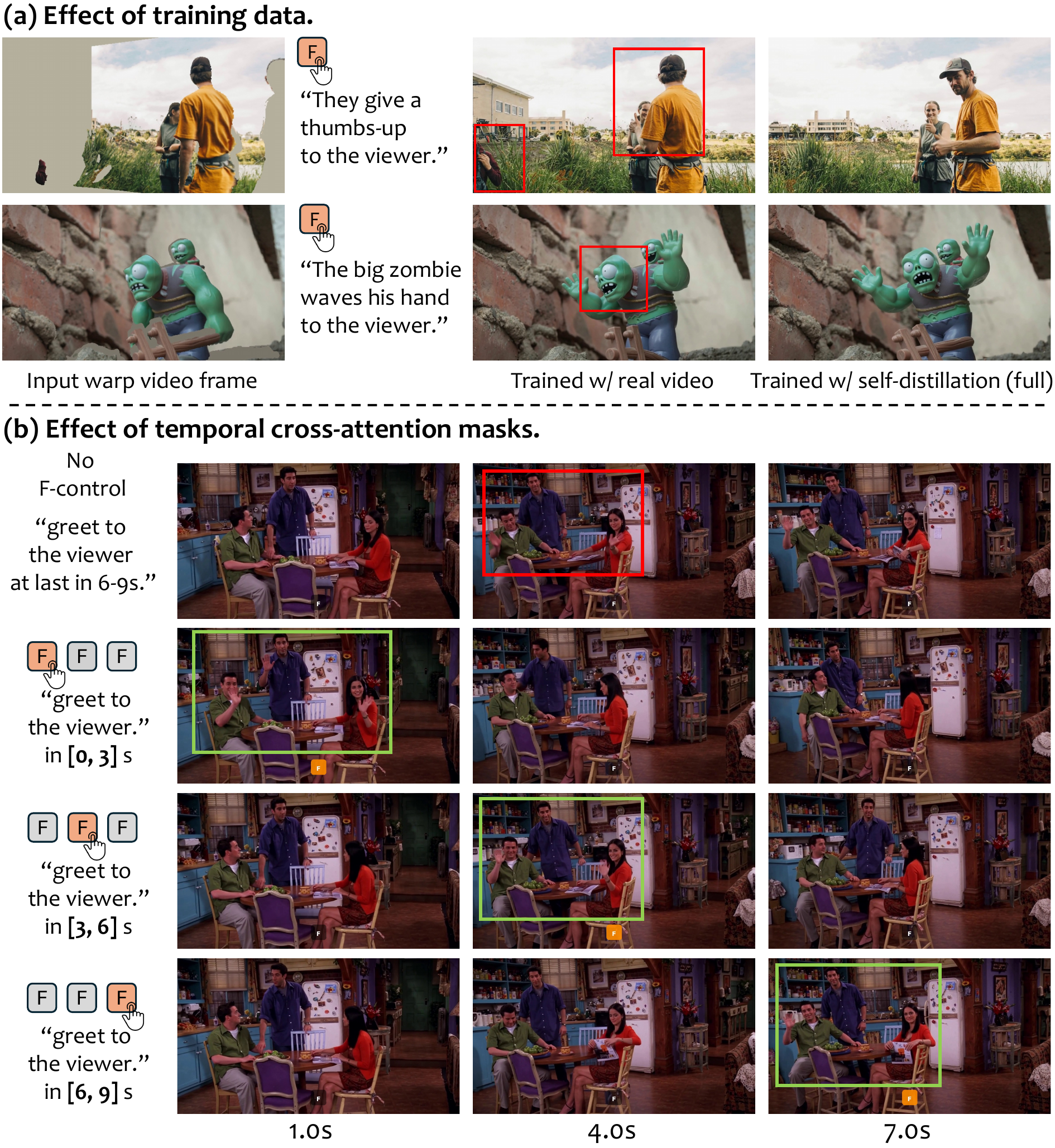}
\caption{Qualitative ablations of HelloWorld. (a) Without interaction data, the model fails to face the viewer and hallucinates (red boxes). (b) Without the temporal mask, the greeting mistimes (red box). With
the mask, it precisely follows the \texttt{F}-press window (green
boxes).}
\label{fig:ablation_selfdis}
\end{figure}

\paragraph{Effect of training data.}
We study the effect of training data by comparing three settings:
\emph{Real-video} trains the LoRA on real data \cite{wang2026warp}, which contains no social interaction; \emph{Human-only} trains on self-generated videos with interactions, but restricted to human characters; and \emph{Full} further extends the coverage to non-human subjects such as animals and cartoon characters.
As shown in Table~\ref{tab:ablation_data}, the self-generated data (Human-only and Full) substantially improves ActAcc and GazeDev, while other metrics are barely affected.
TimeAcc remains unchanged as expected, since the interaction timing is
controlled by the training-free mask rather than training data.
Video quality and camera following are likewise stable, indicating that our self-distillation improves the interaction ability without side effects.

Fig.~\ref{fig:ablation_selfdis} (a) further illustrates the difference: model trained with real videos performs the prompted action, but not toward the viewer.
We attribute this to the training data. Real videos without interactions merely teach the model to inpaint the missing regions of the warp video, providing no signal for engaging with the camera.
In contrast, the self-generated interaction data natively contains characters gazing and acting toward the camera, and the model trained on it thus learns to produce viewer-directed actions.

\begin{table}[t]
\centering
\caption{Ablation on the temporal cross-attention masks.}
\label{tab:ablation_gating}
\small
\setlength{\tabcolsep}{2.6mm}
\begin{tabular}{cc|ccc}
\Xhline{1.0pt}
\rowcolor[gray]{0.92}
$M_v$ & $M_a$ & ActAcc $\uparrow$ & TimeAcc $\uparrow$ & SpeechInWin $\uparrow$ \\
\hline
 & & \textbf{42.5} & 36.7 & 52.5 \\
\checkmark & & 41.5 & 80.9 & 62.8 \\
\checkmark & \checkmark & 41.4 & \textbf{81.7} & \textbf{69.1} \\
\Xhline{1.0pt}
\end{tabular}
\end{table}

\paragraph{Effect of temporal cross-attention masks.}
We compare three settings of the temporal cross-attention mask: no mask, masking the video stream only ($M_v$), and masking both the video and audio streams ($M_v+M_a$).
We additionally report SpeechInWin, which transcribes the generated audio with Whisper~\citep{radford2023robust} and checks whether the speech starts within the interaction window. Its accuracy is computed
in the same way as TimeAcc.
As shown in Table~\ref{tab:ablation_gating}, the mask on each stream improves the temporal control of the corresponding modality, and applying both yields the best overall performance.
Interestingly, the no-mask setting attains the highest ActAcc.
This is expected, as without temporal constraints, the character may perform the prompted action throughout the video, which ActAcc rewards regardless of timing.
The slight drop is thus the cost of temporal localization, rather than a loss of interaction ability.

As shown in Fig.~\ref{fig:ablation_selfdis} (b), without the temporal mask, the interaction still occurs but at an arbitrary time, even when the timing is explicitly specified in the text prompt.
With the mask applied, the interaction precisely falls within the designated window, confirming that our training-free mask equips the world model with temporal control over interactions.

\begin{table}[t]
\centering
\caption{Computational cost of HelloWorld and baselines.}
\label{tab:time}
\small
\resizebox{\columnwidth}{!}{
\setlength{\tabcolsep}{1.2mm}
\begin{tabular}{l|cccc}
\Xhline{1.0pt}
\rowcolor[gray]{0.92}
\bf Method & Time (s) & Time/frame (s) & Res. / frames & FLOPs ($\times10^{15}$) \\
\hline
\color{gray} LTX-2.3 & \color{gray} 50.3 & \color{gray} 0.21 & \color{gray} $1280\times704$ / 241 & \color{gray} 6.9 \\
\hline
WorldPlay & 131.8 & 0.56 & $832\times480$ / 237 & 29.3 \\
Matrix-Game 3.0 & 62.5 & \textbf{0.15} & $1280\times704$ / 417 & 11.2 \\
LingBot-World & 63.3 & 0.39 & $832\times464$ / 153 & 18.2 \\
SANA-WM & \textbf{44.9} & 0.27 & $1280\times704$ / 168 & \textbf{3.2} \\
\hline
\textbf{HelloWorld} & \underline{60.2} & \underline{0.26} & $1280\times704$ / 241 & \underline{9.4} \\
\Xhline{1.0pt}
\end{tabular}
}
\end{table}

\paragraph{Computational cost.}
Table~\ref{tab:time} reports the average computational cost of generating a 10-second clip. Compared with the base model LTX-2.3, the additional warp-video tokens increase the inference time by roughly 20\% (50.3s $\rightarrow$ 60.2s) and the FLOPs by 36\% (6.9 $\rightarrow$ 9.4$\times10^{15}$), which is the modest price of camera controllability.
Despite operating at high resolution (1280$\times$704, 241 frames), HelloWorld remains competitive with the baselines: its per-frame time (0.26\,s) is on par with SANA-WM (0.27\,s) and well below WorldPlay (0.56\,s) and LingBot-World (0.39\,s).

\subsection{User Study}
\label{sec:human}

\begin{table}[t]
\centering
\caption{User study. Each cell reports the percentage of judgments preferring HelloWorld over the competing method, with the bootstrap 95\% confidence interval. $N$ is the number of judgments.}
\label{tab:user}
\small
\resizebox{\columnwidth}{!}{
\setlength{\tabcolsep}{.5mm}
\begin{tabular}{l|c|ccc}
\Xhline{1.0pt}
\rowcolor[gray]{0.92}
\bf HelloWorld \emph{vs.} & $N$ & Action Naturalness & Interaction & Scene Quality \\
\hline
Real-video LoRA & 270 & 83.7 \scriptsize{[79.6, 88.1]} & 88.5 \scriptsize{[84.8, 92.2]} & 82.6 \scriptsize{[78.1, 87.0]} \\
SANA-WM & 330 & 90.9 \scriptsize{[87.6, 93.9]} & 85.5 \scriptsize{[81.8, 89.1]} & 91.2 \scriptsize{[87.9, 94.2]} \\
Warp-as-History & 330 & 81.2 \scriptsize{[77.0, 85.5]} & 80.6 \scriptsize{[76.4, 85.2]} & 83.6 \scriptsize{[79.4, 87.6]} \\
LingBot-World & 300 & 77.3 \scriptsize{[72.7, 82.0]} & 71.7 \scriptsize{[66.7, 77.0]} & 88.0 \scriptsize{[84.0, 91.7]} \\
\Xhline{1.0pt}
\end{tabular}
}
\end{table}

To evaluate the perceptual quality of the interactions, we conduct a user study on 41 samples of HelloWorldBench. We invite 30 human raters, and present them with pairs of videos generated by HelloWorld and by a competing method from the same text prompt and camera motion, in a randomized order. For every pair, the raters answer three questions: (1) \textit{In which video is the intended action performed more naturally?} (2) \textit{In which video does the character feel more like it is interacting with you, the viewer?} and (3) \textit{Which video has higher scene quality?} The results are reported in Table~\ref{tab:user}. HelloWorld is preferred over all four competing methods on all three questions, with every confidence interval lying above 66\%.
This confirms that it achieves state-of-the-art performance in the social interaction dimension as perceived by human viewers.
In particular, the comparison against the LoRA trained on real videos yields the largest margin on the interaction question.
This further verifies that our self-distillation strategy preserves the social prior and substantially improves the quality of the interaction.

\section{Conclusion}
This paper presents HelloWorld, a video world model that enables social interaction with in-world characters. We propose a self-distillation pipeline that turns a base video generation model into an interactive world model without any manually collected or annotated data. We further introduce a temporal cross-attention mask that achieves training-free control over the interaction timing. In addition, we contribute HelloWorldBench, the first benchmark for socially interactive world models. Experiments demonstrate that HelloWorld surpasses existing world models on the interaction metrics while maintaining state-of-the-art video quality.

\textbf{Limitation and future work.} Constrained by the design of the base model and the computation cost, HelloWorld does not yet support real-time interaction with users. World generation is driven by pre-specified camera trajectories and interaction scripts. Our future work will primarily explore autoregressive architectures to enable real-time interaction. Other directions include supporting long video generation and consistent modeling of individual characters, \textit{e.g.}, persistent identities and sustained multi-round interactions.

\bibliographystyle{plainnat}
\bibliography{aaai2027}

\end{document}